\documentclass{article}
\usepackage{times}
\usepackage{graphicx} 
\usepackage{subfigure} 
\usepackage{multirow}
\usepackage[table]{xcolor}
\usepackage{natbib}

\usepackage{algorithm}
\usepackage{algorithmic}
\usepackage{amsmath}
\usepackage{amssymb}
\usepackage{tikz}

\usepackage{hyperref}

\usepackage[arxiv]{icml2017} 

\usepackage{enumitem}
\setenumerate{noitemsep,topsep=0pt,parsep=0pt,partopsep=0pt}

\icmltitlerunning{Variational Inference using Implicit Distributions}
\newcommand{\iid}{i.\,i.\,d.\ }
\newcommand{\eg}{e.\,g.\ }
\newcommand{\ie}{i.\,e.\ }
\newcommand{\kl}[2]{\operatorname{KL}\left[#1\middle\|#2\right]}
\newcommand{\expect}[1]{\mathbb{E}_{#1}}
\newcommand{\priorratio}{r}
\newcommand{\jointratio}{s}
\newcommand{\priordenoiser}{u}

\newcommand{\pdata}{p_{\mathcal{D}}}
\newcommand{\yestick}{\textcolor{green!50!black}{$\checkmark$}}
\newcommand{\nocross}{\textcolor{red!70!black}{}}
\newcommand{\infonly}{\textcolor{blue!70!black}{I}}
\newcommand{\softplus}{\xi^{+}}
\newcommand{\softminus}{\xi^{-}}
\newcommand{\argmin}{\operatorname{argmin}}

\definecolor{Gray}{gray}{0.9}
\begin{document} 

\twocolumn[
\icmltitle{Variational Inference using Implicit Distributions}

\icmlsetsymbol{equal}{*}

\begin{icmlauthorlist}
\icmlauthor{Ferenc Husz\'{a}r}{twtr}
\end{icmlauthorlist}
    
\icmlaffiliation{twtr}{Twitter, London, U.K}

\icmlcorrespondingauthor{Ferenc Husz\'{a}r}{fhuszar@twitter.com}

\icmlkeywords{variational inference, adversarial training, density ratio estimation}

\vskip 0.3in
]

\printAffiliationsAndNotice{}

\begin{abstract} 
Generative adversarial networks (GANs) have given us a great tool to fit implicit generative models to data. Implicit distributions are ones we can sample from easily, and take derivatives of samples with respect to model parameters. These models are highly expressive and we argue they can prove just as useful for variational inference (VI) as they are for generative modelling. Several papers have proposed GAN-like algorithms for inference, however, connections to the theory of VI are not always well understood. This paper provides a unifying review of existing algorithms establishing connections between variational autoencoders, adversarially learned inference, operator VI, GAN-based image reconstruction, and more. Secondly, the paper provides a framework for building new algorithms: depending on the way the variational bound is expressed we introduce prior-contrastive and joint-contrastive methods, and show practical inference algorithms based on either density ratio estimation or denoising.
\end{abstract}

\section{Introduction}
\label{introduction}

Implicit distributions are probability models whose probability density function may be intractable, but there is a way to
\begin{enumerate}
\item sample from them exactly and/or calculate and approximate expectations under them, and
\item calculate or estimate gradients of such expectations with respect to model parameters.
\end{enumerate}

A popular example of implicit models are stochastic generative networks: samples from a simple distribution - such as uniform or Gaussian - are transformed nonlinearly and non-invertably by a deep neural network. Such networks can flexibly parametrise a wide range of probability distributions, including even degenerate ones which may not even have a continuous density.

Implicit models have been successfully applied to generative modelling in generative adversarial networks \citep[GANs][]{goodfellow2014generative} and subsequent work \citep{salimans2016improved, radford2015unsupervised, donahue2017adversarial, dumoulin2017adversarially}. They work particularly well for visual data, partly because they can exploit the inductive biases of convolutional neural networks, and partly because they can flexibly model potentially degenerate, manifold-like distributions which natural images are assumed to follow.

This note is about using implicit distributions in another important probabilistic machine learning problem: approximate inference in latent variable models. Unlike in the first applications of GANs to generative modelling where an implicit model directly models the distribution of observed data, in approximate inference we are interested in modelling the posterior distribution of latent variables given observed data. Direct generative modelling and approximate inference are very different problems indeed: in the former we are provided with samples $x_i$ from the distribution to be modelled, in the latter we are given a joint distribution of latents $z$ and observations $x$, and a set of observed samples $x_i$, but no samples from the posterior itself.

In this note we focus on variational inference (VI) which works by minimising a divergence between the approximate and real posteriors. More precisely we follow the usual KL-divergence formulation, but other, more general variational methods exist \citep[\eg][]{li2016renyi,ranganath2016operator}. VI also provides a lower bound to the marginal likelihood or model evidence - the \emph{evidence lower bound} or ELBO - which can be maximised with respect to parameters of the latent variable model to approximate maximum likelihood learning. It's important to keep in mind that despite several algorithms in this note look and feel like adversarial training procedures, the way the model is fit to observed data is more akin to variational auto-encoders \citep[VAE,][]{kingma2015autoencoding,icml2015_rezende15} than to GANs.

There are several reasons to explore implicit distributions in the context of variational inference. Firstly, explicit VI  is often limited to exponential family distributions or other distributions with tractable densities \citep{icml2015_rezende15,kingma2016improving} which may not be expressive enough to capture non-trivial dependencies that the real posterior exhibits. A flexible, implicit distribution may provide a better approximation to the posterior and a sharper lower bound. Secondly, It may be desirable to use an implicit likelihood as the resulting latent variable model may fit the data better. For example, the likelihood or forward model might be described as a probabilistic program \citep{vajda2014probabilistic} whose density is intractable or unknown. Finally, sometimes we may want to use implicit priors over latent variables. For example in a deep hierarchical latent variable model the prior for a layer may be a complicated probabilistic model with an intractable density. Or, when solving inference problems in computational photography, the prior may be the empirical distribution of natural images as in \citep{sonderby2016amortised}. In summary, any or all of the prior, the likelihood and the approximate posterior may have to be modelled implicitly, and we need VI  procedures that are ready to tackle these situations.

In this note we present two sets of tools to handle implicit distributions in variational inference: GAN-like \emph{adversarial algorithms} which rely on density ratio estimation, and \emph{denoising-based algorithms} which build a representation of the gradients of each implicit distribution's log-density and use these gradient estimates directly in a stochastic gradient descent (SGD) algorithm. We further classify algorithms as \emph{prior-contrastive} and \emph{joint-contrastive} depending on which form of the variational bound they use. Prior-contrastive methods only deal with implicit distributions over latent variables (\ie the prior or approximate posterior), while joint-contrastive methods can handle fully implicit models where none of the distributions involved has a tractable density. This classification gives rise to a range of algorithms listed in Table \ref{summary-table}, alongside related algorithms from prior work. All of the algorithms presented here can perform variational approximate inference, which is the main focus of this note, but not all of them can perform learning unless the likelihood is explicitly defined.

\setlength{\tabcolsep}{3pt} 
\begin{table}[t]

\label{summary-table}
\vskip 0.15in
\begin{center}
\begin{small}
\begin{sc}
\begin{tabular}{|l|ccc|r|}
\hline
& \multicolumn{3}{c|}{implicit} & \\
\multirow{-2}{*}{Algorithm} & \tiny $p_{\theta}(z)$ & \tiny $p_\theta(x\vert z)$ & \tiny $q_\psi(z\vert x)$ & \multirow{-2}{*}{VI}\\
\hline\hline
VAE\ {\tiny \citep{kingma2015autoencoding}} & & & & \\
NF\ {\tiny \citep{icml2015_rezende15}} & & & & \multirow{-2}{*}{\yestick} \\
\rowcolor{gray!20}PC-Adv, {\tiny Algorithm\ \ref{alg:PCAdv} }            & & & & \\
\rowcolor{gray!20}AffGAN$^\dagger$\ {\tiny \citep{sonderby2016amortised}}  & & & & \\
\rowcolor{gray!20}AVB\ {\tiny \citep{mescheder2017adversarial}}  & \multirow{-3}{*}{\infonly} & \multirow{-3}{*}{\nocross} & \multirow{-3}{*}{\yestick} & \multirow{-3}{*}{\yestick} \\
OPVI\ {\tiny \citep{ranganath2016operator}} & \infonly & \nocross & \yestick & \yestick \\
\rowcolor{gray!20}PC-Den,\ {\tiny Algorithm\ \ref{alg:PCDen}}                 & \infonly & \nocross & \yestick & \yestick \\
JC-Adv,\ {\tiny Algorithm\ \ref{alg:JCAdv}}                & \infonly & \infonly & \yestick & \yestick \\
\rowcolor{gray!20}JC-Den                 & \infonly & \infonly & \yestick & \yestick \\
JC-Adv-RMD$^\ddagger$            & \yestick & \yestick & \yestick & \yestick \\
\hline
\rowcolor{gray!20}AAE\ {\tiny \citep{makhzani2016adversarial}}   & \infonly & \nocross & \yestick & \nocross \\
DeePSiM\ {\tiny \citep{dosovitskiy2016generating}} & \infonly & \nocross & \yestick & \nocross\\
\rowcolor{gray!20}ALI\ {\tiny \citep{dumoulin2017adversarially}} & & & & \\
\rowcolor{gray!20}BiGan\ {\tiny \citep{donahue2017adversarial}} & \multirow{-2}{*}{\yestick} & \multirow{-2}{*}{\yestick} & \multirow{-2}{*}{\yestick} & \multirow{-2}{*}{\nocross}\\

\hline
\end{tabular}
\end{sc}
\end{small}
\end{center}
\caption{Summary of algorithms for variational inference and learning in latent variable models using implicit distributions. Columns 2-4 indicate whether the component distributions: the prior, likelihood or approximate posterior are handled implicitly by the algorithm. ``\infonly'' denotes inference only - the parameters $\theta$ of these distributions cannot be learned unless they are explicitly defined, but inference can still be performed. The last column indicates whether the algorithm has a variational interpretation, \ie minimises an approximation to ELBO. In naming the algorithms \texttt{PC} and \texttt{JC} stand for prior-contrastive or joint-contrastive, \texttt{Adv} and \texttt{Den} stand for adversarial (Section \ref{sec:ratioestimation}) or denoiser-based (Section \ref{sec:denoiser_guided}). Algorithms that share a row are equivalent or special cases of each other. $^\dagger$\texttt{AffGAN} is specialised to the task of image super-resolution where the likelihood is degenerate and linear. $^\ddagger$ The reverse-mode differentiation-based \texttt{JC-Adv-RMD} algorithm has not been validated experimentally.}
\end{table}

\subsection{Overview of prior work}

Several of the algorithms proposed here have been discovered in some form before. However, their connections to variational inference is rarely made explicit. In this section we review algorithms for inference and feature learning which use implicit distributions or adversarial training. As we will see, several of these admit a variational interpretation or can be rather straightforwardly modified to fit the variational framework.

GANs have been used rather successfully to solve inverse problems in computer vision. These inverse problems can be cast as a special case of approximate inference. \citet{dosovitskiy2016generating} used GANs to reconstruct and generate images from non-linear feature representations. As pointed out later by \citet{sonderby2016amortised}, this method, DeePSiM, can be interpreted as a special case of amortised maximum a posteriori (MAP) or variational inference with a Gaussian observation model. GANs have also been used for inference in image super-resolution \citet{ledig2016photo, sonderby2016amortised}. Connections between GANs and VI  in this context were first pointed out in \citep[][Appendix F]{sonderby2016amortised}. \citet{sonderby2016amortised} also introduced a modified objective function for the GAN generator which ensures that the algorithm minimises the Kullback-Leibler divergence as opposed to the Jensen-Shannon divergence, an essential step to using GANs for VI. The AffGAN algorithm presented there is highly application specific, thus, it does not solve VI in general.

In more recent, parallel work, \citep{mescheder2017adversarial} proposes \emph{adversarial variational Bayes}(AVB), perhaps the best description of the use of GANs for variational inference. AVB is a general algorithm that allows for implicit variational distributions and is in fact equivalent to the prior-contrastive adversarial algorithm (\texttt{PC-Adv}, Algorithm \ref{alg:PCAdv}) described in Section \ref{sec:ratioestimation}. Operator variational inference \citep[OPVI,][]{ranganath2016operator} formulates a general class of variational lower bounds based on operator divergences, resulting in a practical algorithm for training implicit inference networks without a tractable density. As is shown in the paper, the KL-divergence-based variational bound used here and in \citep{mescheder2017adversarial} is a special case of OPVI. Adversarial autoencoders \citep[AAE,][]{makhzani2016adversarial} are similar to variational autoencoders where the KL-divergence term is replaced by an adversarial objective. However, AAEs do not use the KL-divergence formulation of the adversarial loss and their discriminator is independent of the encoder's input, thus they are not a true variational method. Finally, \citet{karaletsos2016adversarial} proposed variational message passing, in which adversaries are employed to minimise local Jensen-Shannon divergences in an algorithm more akin to expectation propagation \citep{minka2001expectation} than to variational inference.

Another line of research extends GANs to latent variable models by training the discriminator on the joint distribution of latent and observed variables. This technique has been independently discovered as bi-directional GAN \citep[BiGAN,][]{dumoulin2017adversarially} and adversarially learned inference \citep[ALI,][]{donahue2017adversarial}. These algorithms are closely related to the joint-contrastive adversarial algorithm (\texttt{JC-Adv}, Algorithm \ref{alg:JCAdv}). ALI and BiGAN use the Jensen-Shannon formulation of GANs rather than the KL-divergence ones used here. On the one hand, this means that the Jensen-Shannon variants aren't technically VI  algorithms. On the other hand, the symmetry of the Jensen-Shannon divergence makes ALI and BiGAN completely symmetric, enabling not only approximate inference but also learning in the same algorithm. Unfortunately, this is no longer true when KL divergences are used: \texttt{JC-Adv} is an algorithm for inference only.

The algorithms mentioned so far are examples of adversarial techniques which rely on density ratio estimation as the primary tool for dealing with implicit distributions \citep{mohamed2016learning}. \citet{sonderby2016amortised} and \citet{wardefarley2017improving} demonstrated an alternative or complementary technique based on denoising autoencoders. As shown by \citep{alain2014regularized} the optimal denoising function learns to represent gradients of the log data density - which in turn can be used in an inference method. \citet{sonderby2016amortised} used this insight to build a denoiser-based inference algorithm for image super-resolution and connected it to amortised maximum a posteriori (MAP) inference. The extension from MAP to variational inference is straightforward and this method is closely related to the prior-contrastive denoising VI (\texttt{PCDen}, Algorithm \ref{alg:PCDen}) algorithm presented here.


\section{Variational Inference: Two forms}\label{sec:VI_formulations}

In this section, we give a lightweight overview of amortised variational inference (VI) in a latent variable model, in a model similar to \eg variational autoencoders \citep{kingma2015autoencoding}. We observe an \iid sequence of $N$ observed data $\mathcal{D} = \{x_n, n=1\ldots N\}$. For each data point there exists an associated latent variable $z_n, n=1\ldots N$. We specify a prior $p_\theta(z)$ over latent variables and a forward model $p_\theta(x\vert y)$ which describes how the observations are related to latents. In such model we are interested in maximum likelihood learning, which maximises the marginal likelihood or model evidence $\sum_{n=1}^{N} \log p_\theta(x_n)$ with respect to parameters $\theta$, and inference which involves calculating the posterior $p_\theta(z\vert x)$. We assume that neither the marginal likelihood or the posterior are tractable.

In amortized VI we introduce an auxiliary probability distribution $q(z\vert x_n; \psi)$, known as the recognition model, inference network or approximate posterior. Using $q_\psi$ we define the evidence lower bound (ELBO) as follows:
\begin{equation}
\mathcal{L}(\theta, \psi) = \sum_{n=1}^{N} \left\{ \log p_\theta(x_n) - \kl{q_\psi(z\vert x_n)}{p_\theta(z\vert x_n)} \right\}
\end{equation}

As the name suggests, ELBO is a lower bound to the model evidence $p_\theta(\mathcal{D})$ and it is exact when $q_\psi$ matches the true posterior $p_\theta(z\vert x)$ exactly. Maximising ELBO with respect to $\psi$ is known as \emph{variational inference}. This minimises the KL divergence $\kl[q_\psi \vert p_\theta(z\vert x)]$ thus moving the $q_\psi$ closer to the posterior. Conversely, maximising ELBO with respect to $\theta$ is known as \emph{variational learning} which approximates maximum likelihood learning.

The ELBO can be calculated exactly for many combinations of $p_\theta$ and $q_\psi$, whose densities are tractable. VAEs use a re-parametrisation trick to construct a low variance estimator to ELBO, but still require tractable densities for both the model $p_\theta$ and recognition model $q_\psi$. If $p_\theta$ and/or $q_\psi$ are implicit the ELBO needs to be approximated differently. As we will see in the next sections, it is useful to formulate ELBO in terms of density ratios.

There are two main forms considered here. Firstly, the \emph{prior-contrastive} form used also by VAEs \citep{kingma2015autoencoding}:
\begin{align}
&\mathcal{L} = \sum_{n=1}^{N}\expect{z\sim q_\psi(z\vert x_n)} \left[ \log p_\theta(x_n\vert z) - \priorratio_{\theta, \psi}(z,x_n) \right]\label{eqn:priorcontrastive}\\
&= \sum_{n=1}^{N}\left[\expect{z\sim q_\psi(z\vert x_n)}\log p_\theta(x_n\vert z) - \kl{q_\psi(z\vert x_n)}{p_\theta(z)}\right],\notag
\end{align}
where we introduced notation for the logarithmic density ratio $\priorratio_{\theta, \psi} = \log\frac{q_\psi(z\vert x_n)}{p_\theta(z)}$.

We call Eqn.\ \eqref{eqn:priorcontrastive} the \emph{prior-contrastive} expression as the $KL$ term contrasts the approximate posterior $q_\psi$ with the prior $p_\theta(z)$.Alternatively, we can write ELBO in a \emph{joint-contrastive} form as follows:

\begin{align}
&\mathcal{L}(\theta, \psi) = -N\cdot\left(\kl{q_\psi(z\vert x)\pdata(x)}{p_\theta(x, z)} - \mathbb{H}[\pdata]\right)\notag\\
&= - \sum_{n=1}^{N}\expect{z\sim q_\psi(z\vert x_n)}\jointratio_{\theta, \psi}(x_n, z) - N\cdot\mathbb{H}[\pdata],\label{eqn:jointcontrastive}
\end{align}


where we introduced notation $\pdata$ to denote the real data distribution and $\mathbb{H}[\pdata]$ denotes its entropy\footnote{In practice, $\pdata$ is an empirical distribution of samples, so technically it does not have a continuous density or differential entropy $\mathbb{H}[\pdata]$. We still use this notation liberally to avoid unnecessarily complicating the derivations.}. $\mathbb{H}[\pdata]$ can be ignored as it is constant with respect to both $\theta$ and $\psi$. We also introduced notation $\jointratio_{\theta, \psi}$ to denote the logarithmic density ratio $\jointratio_{\theta, \psi}(x_n, z) = \log \frac{q_\psi(z\vert x_n)\pdata(x)}{p_\theta(x_n, z)}$. Note that while $\priorratio_{\theta, \psi}$ was a log-ratio between densities over $z$, $\jointratio_{\theta, \psi}$ is the ratio of joint densities over the tuple $(x,z)$. As this form contrasts joint distributions, we call Eqn.\ \eqref{eqn:jointcontrastive} the \emph{joint-contrastive} expression.

When using implicit models the density ratios $\priorratio_{\theta, \psi}$ and $\jointratio_{\theta, \psi}$ cannot be computed analytically. Indeed, even if all distributions involved are explicitly defined, $\pdata$ is only available as an empirical distribution, thus $\jointratio_{\theta, \psi}$ cannot be calculated even if the densities of other distributions are tractable. In this note we rely on techniques for estimating $\priorratio_{\theta, \psi}$ or $\jointratio_{\theta, \psi}$, or their gradients, directly from samples. For this to work we need to deal with a final difficulty: that $\priorratio_{\theta, \psi}$ or $\jointratio_{\theta, \psi}$ themselves implicitly depend on the parameter $\psi$ which we would like to optimise.

\subsection{Dependence of $\priorratio_{\theta, \psi}$ and $\jointratio_{\theta, \psi}$ on $\psi$}

The KL-divergences in equations \label{eqn:priorcontrastive} and \label{eqn:jointcontrastive} depend on $\psi$ in two ways: first, an expectation is taken with respect to $q_\psi$ - this is fine as we assumed expectations under implicit distributions and their gradients can be approximated easily. Secondly, the ratios $\priorratio_{\theta, \psi}$ and $\jointratio_{\theta, \psi}$ themselves depend on $\psi$, which may cause difficulties. If one optimised ELBO na\:{i}vely via gradient descent, one should back-propagate through both of these dependencies. Fortunately, the second dependence can be ignored: 

\begin{equation}
\frac{\partial}{\partial \psi} \left.\expect{z \sim q_\psi} \priorratio_{\theta, \psi}(z) \right\vert_{\psi = \psi_0} = \left. \frac{\partial}{\partial \psi} \expect{z \sim q_\psi} \priorratio_{\theta, \psi_0}(z) \right\vert_{\psi = \psi_0}\label{eqn:implicit_dependency_ignored}
\end{equation}

The only difference between the LHS and RHS of the equation is in the subscripts $\priorratio_{\theta, \psi}$ v.\,s.\ $\priorratio_{\theta, \psi_0}$. As $\priorratio_{\theta, \psi_0}$ is a constant with respect to $\psi$, Eqn.\ \eqref{eqn:implicit_dependency_ignored} reduces to the gradient of an expectation with respect to $q_\psi$, which we assumed we can approximate if $q_\psi$ is an implicit distribution. The detailed proof of Eqn.\ \eqref{eqn:implicit_dependency_ignored} is in Appendix \ref{sec:proof_implicit_dependency}, the key idea of which is the observation that for any $\psi_0$
\begin{equation}
\expect{z \sim q_\psi} \priorratio_{\theta, \psi}(z) = \expect{x, z \sim q_{\psi_0}}  \priorratio_{\theta, \psi}(x, z) + \kl{q_\psi}{q_{\psi_0}}\notag
\end{equation}

A similar equation analogously holds for $\jointratio_{\theta, \psi}$ in Eqn.\ \eqref{eqn:jointcontrastive}, or indeed, any other KL divergence as well.

\subsection{Approximate SGD Algorithms for VI}

In the following sections we outline algorithms for VI which allow for implicit distributions. These algorithms can generally described as two nested loops of the following nature:

\begin{itemize}
\item the outer loop performs stochastic gradient descent (SGD) on an approximation to ELBO with respect to $\phi$, using gradient estimates obtained by the inner loop
\item in each iteration of outer loop, with $\psi=\psi_t$ fixed, the inner loop constructs an estimate to $\priorratio_{\theta, \psi_t}$, $\jointratio{\theta, \psi_t}$, or more generally to the gradient in Eqn.\ \eqref{eqn:implicit_dependency_ignored}
\end{itemize}

As long as the gradient estimates provided by the inner loop has no memory between subsequent iterations of the outer loop, and the gradient estimates provided by the inner loop on average constitute a conservative vector field, the algorithms can be seen as instances of SGD, and as such, should have the convergence properties of SGD.


\section{Direct Density Ratio Estimation}\label{sec:ratioestimation}

Direct density ratio estimation, also known as direct importance estimation, is the task of estimating the ratio between the densities of two probability distribution given only i.\,i.\,d.\ samples from each of the distributions \citep[see \eg]{kanamori2009least, sugiyama2008kliep, mohamed2016learning}. This task is relevant in many machine learning applications, such as dealing with covariate shift or domain adaptation. A range of methods have been introduced to learn density ratios from samples, here we focus on \emph{adversarial} techniques which employ a discriminator trained via logistic regression. We note that other methods such as KLIEP \citep{sugiyama2008kliep, mohamed2016learning} or LSIF \citep{kanamori2009least, uehara2016generative} could be used just as well.

\subsection{Adversarial approach using discriminators}

\citep{bickel2007discriminative} proposed estimating density ratios by training a logistic regression classifier between samples from the two distributions. Assuming the classifier is close to a unique Bayes-optimum, it can then be used directly to provide an estimate of the logarithmic density ratio. This approach has found great application in generative adversarial networks \citep{sonderby2016amortised, mohamed2016learning, uehara2016generative}, which work particularly well for generative modelling of images \citep[see \eg][]{salimans2016improved}.

Let us use this to construct an approximation $\priorratio_\phi$ to the logarithmic density ratio $\priorratio_{\psi,\theta}$ from Eqn.\ \eqref{eqn:priorcontrastive}. We can do this by minimising the following objective function, typically via SGD:

\begin{align}
\ell^{\tiny \texttt{PC-Adv}}_{\psi,\theta}(\phi) &= \sum_{n=1}^{N} \expect{z\sim p_\theta} \softplus \left(\priorratio_\phi(x_n,z)\right) \notag\\
    &- \sum_{n=1}^{N} \expect{z\sim q_\psi(z\vert x_n)} \softminus \left(\priorratio_\phi(x_n,z)\right), \label{eqn:JC_discriminator_loss}
\end{align}

where $\softplus(t) = \log(1+\exp^{t})$ and $\softminus(t) = t - \softplus(t)$ are the softplus and softminus functions, respectively. Once the approximate log ratio $\priorratio_\phi$ is found, we can use it to take a gradient descent step along the approximate negative ELBO:

\begin{align}
-\mathcal{L}_{\phi}(\psi, \theta) &= \sum_{n=1}^{N} \expect{\epsilon\sim\mathcal{N}(0,I)} \priorratio_{\phi}\left(x_n,g_\psi(x_n,\epsilon)\right)\notag\\
    &- \sum_{n=1}^{N} \expect{\epsilon\sim\mathcal{N}(0,I)} \log p_\theta\left(x_n\vert g_\psi(x_n,\epsilon)\right), \label{eqn:JC_generator_loss}
\end{align}

where we re-parametrised sampling from $q_\psi$ in terms of a generator function $g_\psi$ and noise $\epsilon$. When $q_\psi$ is explicitly defined, this re-parametrisation is the same as the re-parametrisation in VAEs. When $q_\psi$ is an implicit distribution, it often already defined in terms of a non-linear function $g_\psi$ and a noise variable $\epsilon$ which it transforms.

Equations \label{eqn:JC_discriminator_loss} and \label{eqn:JC_generator_loss} are analogous to the discriminator and generator losses in generative adversarial networks, with $1/(1+\exp(-\priorratio_\phi))$ taking the role of the discriminator. Optimising the two losses in tandem gives rise to the prior-contrastive adversarial algorithm (\texttt{PC-Adv}, Algorithm \ref{alg:PCAdv}) for variational inference. This algorithm is equivalent to the independently developed \emph{adversarial variational Bayes} \citep[][AVB]{mescheder2017adversarial}.

As the likelihood $p_\theta(x\vert z)$ appears in Eqn.\ \eqref{eqn:JC_generator_loss}, in Algorithm \ref{alg:PCAdv} the forward model has to be explicitly defined, but the prior $p_\theta(z)$ and approximate posterior $q_\psi$ can be implicit. Algorithm \ref{alg:PCAdv} only describes variational inference - finding $\psi$ given $\theta$ - but the approximate ELBO in Eqn.\ \label{eqn:JC_generator_loss} can be used for variational learning of $\theta$ as well, with the exception for parameters of the prior $p_\theta(z)$.

Learning prior parameters involves minimising the KL-divergence $\kl{\sum_{n=1}^{N}q_\psi(z\vert x_n)}{p_\theta(z)}$ which is akin to fitting $p_\theta$ to samples from the aggregate posterior $\sum_{n=1}^{N}q_\psi(z\vert x_n)$ via maximum likelihood. If the prior has a tractable density, this may be an easy task to do. A more interesting case is though when the prior $p_\theta(z)$ itself is a latent variable model, in which case we can lower bound the said KL divergence with another ELBO, thereby stacking multiple models on top of each other in a hierarchical fashion \cite{kingma2015autoencoding, icml2015_rezende15, sonderby2016ladder}.

A similar adversarial algorithm (\texttt{JC-Adv}, Algorithm \ref{alg:JCAdv}) can be constructed to target $\jointratio_{\theta,\psi}$ in the joint-contrastive formulation of ELBO (Eqn.\ \ref{eqn:jointcontrastive}). \texttt{JC-Adv} is very similar to ALI \citep{dumoulin2017adversarially} and BiGAN \citep{donahue2017adversarial} in that it learns to discriminate between the joint distributions $\pdata(x)q_\psi(z\vert x)$ and $p_\theta(x,z)$. Unlike these methods, however, \texttt{JC-Adv} uses the correct loss functions so it maximises an approximation to the ELBO. Unlike in \texttt{PC-Adv}, which required a tractable likelihood $p_\theta(x\vert z)$, \texttt{JC-Adv} also works with completely implicitly defined models. As a downside, \texttt{JC-Adv} provides no direct way for variational learning of $\theta$. ALI and BiGAN exploit the symmetry of the Jensen-Shannon divergence to optimise for $\theta$, but as \texttt{JC-Adv} uses the asymmetric KL-divergence, this is not an option. Section \ref{sec:discussion} explores an idea for fixing this drawback of \texttt{JC-Adv}.

\begin{algorithm}[bt]
   \caption{\texttt{PC-Adv}: prior-contrastive adversarial VI}
   \label{alg:PCAdv}
\begin{algorithmic}
    \STATE {\bfseries Input:} data $\mathcal{D}$, model $p_\theta$, batchsize $B$, iter.\ count $K$
    \REPEAT
        \FOR{$k=1$ {\bfseries to} $K$}
            \STATE $\{z^{p}_{k,b}, b=1 \ldots B\} \leftarrow$ sample $B$ items from $p_\theta$
            \STATE $\{x_{k,b}, b=1 \ldots B\} \leftarrow$ sample $B$ items from $\mathcal{D}$
            \FORALL{$x_{k,b}$}
                \STATE $z^{q}_{k,b} \leftarrow$ sample from $q_\psi(z\vert x_{k,b})$
            \ENDFOR
            \STATE {\bfseries update} $\phi$ by gradient descent step on
            {\footnotesize \vspace{-8pt}
            $$
                \sum_{b=1}^{B} \left[ \softplus\left(\priorratio_\phi(x_{k,b},z^{p}_{k,b})\right) - \softminus\left(\priorratio_\phi(x_{k,b},z^{q}_{k,b})\right) \right]
            $$
            \vspace{-12pt}
            }
        \ENDFOR
        \STATE $\{x_{b}, b=1 \ldots B\} \leftarrow$ sample $B$ items from $\mathcal{D}$
        \STATE $\{\epsilon_{b}, b=1 \ldots B\} \leftarrow$ sample $B$ items from $\mathcal{N}(0,I)$
        \STATE {\bfseries update} $\psi$ by gradient descent step on
        {\footnotesize \vspace{-8pt}
            $$
              \sum_{b=1}^{B} \priorratio_\phi(x_{b},g_\psi(x_{b},\epsilon_{b})) - \log p_\theta(x_{n,b}\vert g_\psi(x_{b},\epsilon_{b}))
            $$
        \vspace{-12pt}
        }
    \UNTIL{change in $\psi$ is negligible}
\end{algorithmic}
\end{algorithm}

\begin{algorithm}[bt]
   \caption{\texttt{JC-Adv}: joint-contrastive adversarial VI}
   \label{alg:JCAdv}
\begin{algorithmic}
    \STATE {\bfseries Input:} data $\mathcal{D}$, model $p_\theta$, batchsize $B$, iter.\ count $K$
    \REPEAT
        \FOR{$k=1$ {\bfseries to} $K$}
            \STATE $\{x^{q}_{k,b}, b=1 \ldots B\} \leftarrow$ sample $B$ items from $\mathcal{D}$
            \FORALL{$x^{q}_{k,b}$}
                \STATE $z^{q}_{k,b} \leftarrow$ sample from $q_\psi(z\vert x^{q}_{k,b})$
            \ENDFOR
            \STATE $\{z^{p}_{k,b}, b=1 \ldots B\} \leftarrow$ sample $B$ items from $p_\theta$
            \FORALL{$z^{p}_{k,b}$}
                \STATE $x^{p}_{k,b} \leftarrow$ sample from $p_\theta(x\vert z^{p}_{k,b})$
            \ENDFOR
            \STATE {\bfseries update} $\phi$ by gradient descent step on
            {\footnotesize \vspace{-8pt}
            $$
                \sum_{b=1}^{B} \left[ \softplus\left(\jointratio_\phi(x^{p}_{k,b},z^{p}_{k,b})\right) - \softminus\left(\jointratio_\phi(x^{q}_{k,b},z^{q}_{k,b})\right) \right]
            $$
            \vspace{-12pt}
            }
        \ENDFOR
        \STATE $\{x_{b}, b=1 \ldots B\} \leftarrow$ sample $B$ items from $\mathcal{D}$
        \STATE $\{\epsilon_{b}, b=1 \ldots B\} \leftarrow$ sample $B$ items from $\mathcal{N}(0,I)$
        \STATE {\bfseries update} $\psi$ by gradient descent step on
        {\footnotesize \vspace{-8pt}
            $$
              \sum_{b=1}^{B} \jointratio_\phi(x_{b},g_\psi(x_{b},\epsilon_{b}))
            $$
        \vspace{-12pt}
        }
    \UNTIL{change in $\psi$ is negligible}
\end{algorithmic}
\end{algorithm}

\begin{algorithm}[tb]
   \caption{\texttt{PC-Den}: prior-contrastive denoising VI}
   \label{alg:PCDen}
\begin{algorithmic}
    \STATE {\bfseries Input:} data $\mathcal{D}$, $p_\theta$, batchsize $B$, iter.\ count $K$, $\sigma_n$
    \REPEAT
        \FOR{$k=1$ {\bfseries to} $K$}
            \STATE $\{x_{k,b}, b=1 \ldots B\} \leftarrow$ $B$ samples from $\mathcal{D}$
            \FORALL{$x_{k,b}$}
                \STATE $z_{k,b} \leftarrow$ sample from $q_\psi(z\vert x_{k,b})$
            \ENDFOR
            \STATE $\eta_{k,b}, b=1 \ldots B\} \leftarrow$ $B$ samples from $\mathcal{N}(0,\sigma_nI)$
            \STATE {\bfseries update} $\phi$ by gradient descent step on
            {\footnotesize \vspace{-8pt}
            $$
                \sum_{b=1}^{B} \| \priordenoiser_\phi(z_{k,b} + \eta_{k,b}) - z_{k,b} \|^2
            $$
            \vspace{-12pt}
            }
        \ENDFOR
        \STATE $\{x_{b}, b=1 \ldots B\} \leftarrow$ sample $B$ items from $\mathcal{D}$
        \STATE $\{\epsilon_{b}, b=1 \ldots B\} \leftarrow$ sample $B$ items from $\mathcal{N}(0,I)$
        \FORALL{$x_{b}$}
            \STATE $z_b \leftarrow g_\psi(x_{b}, \epsilon_b)$
        \ENDFOR
        \STATE {\bfseries update} $\psi$ by gradient descent using gradient
        {\footnotesize \vspace{-8pt}
            $$
              \sum_{b=1}^{B} \frac{\partial g_\psi(x_{b}, \epsilon_b)}{\partial \psi} \left[ \left.\frac{\partial p_\theta(x_b, z)}{\partial z}\right\vert_{z=z_b} + \frac{z_b - \priordenoiser_\phi(z_b)}{\sigma_n^2} \right]
            $$
        \vspace{-12pt}
        }
    \UNTIL{change in $\psi$ is negligible}
\end{algorithmic}
\end{algorithm}



\section{Denoiser-guided learning}\label{sec:denoiser_guided}

Although most versions of GAN use an adversarial discriminator based on logistic regression, there are other ways one can tackle learning and inference with implicit distributions. One interesting tool that has emerged in recent papers \citep{sonderby2016amortised,wardefarley2017improving} is the use of denoising autoencoders \citep[DAEs,\,][]{vincent2008extracting} or reconstruction-contractive autoencoders \citep[RCAEs,\,][]{alain2014regularized}.

The key observation for using denoising is that the Bayes-optimal denoiser function captures gradients of the log density of the data generating distribution:

\begin{align}
v^{*}(x,z) &= \argmin_v \expect{x,z\sim q(x,z), \eta \sim \mathcal{N}_{\sigma_n}} \|v(x,z+\eta) - z\|^2 \notag\\
    &\approx z + \sigma^2_n \frac{\partial \log q(z\vert x)}{\partial z} \text{\hspace{30pt} as } \sigma_n\rightarrow 0
\end{align}

This allows us to construct an estimator to the \emph{score} of a distribution by fitting a DAE to samples. We note that it is possible to obtain a more precise analytical expression for for the optimal denoising function \citep{valpola2016denoising}.

Let's see how one can use this in the prior-contrastive scenario to deal with an implicit $q_\theta$. First, we fit a denoising function $\priordenoiser_\phi$ by minimising the following loss:
\begin{equation}
\ell^{\tiny \texttt{PC-Den}}_{\psi, \sigma_n}(\phi) = \sum_{n=1}^N \expect{z\sim q_\theta(z\vert x_n), \eta\sim \mathcal{N}_{\sigma_n}} \| v_\phi(z + \eta,x_n) - z \|^2,
\end{equation}
which can then be used to approximate the gradient of ELBO (Eqn.\ \eqref{eqn:implicit_dependency_ignored}) with respect to $\psi$ as follows:
{\small
\begin{align}
&\frac{\partial \mathcal{L}(\theta,\psi)}{\partial \psi} \approx \sum_{n=1}^{N} \expect{\epsilon \sim \mathcal{N}} \frac{\partial g_\psi(x_n,\epsilon))}{\partial \psi}  \left.\frac{\partial \log p_\theta(x_n, z)}{\partial z}\right\vert_{z=g_\psi(x_n,\epsilon)}\notag\\
&+ \sum_{n=1}^{N} \expect{\epsilon\sim \mathcal{N}} \frac{\partial g_\psi(x_n,\epsilon))}{\partial \psi} \frac{g_\psi(x_n,\epsilon) - \priordenoiser_\phi(g_\psi(x_n,\epsilon))}{\sigma^2_n}.
\end{align}
}

Several SGD methods only require gradients of the objective function as input, this gradient estimate can be readily used to optimise an approximate ELBO. The resulting iterative algorithm, prior-contrastive denoising VI (\texttt{PC-Den}, Algorithm \ref{alg:PCDen}) updates the denoiser and the variational distribution in tandem. Following similar derivation one can construct several other variants of the algorithm. The denoiser approach is more flexible than the adversarial approach as one can pick and choose which individual distributions are modelled explitly, and which ones are implicit. For example, when the prior $p_\theta$ is implicit, we can train a denoiser to represent its score function. Or, one can start from the joint-contrastive formulation of the ELBO and train a DAE over joint distribution of $x$ and $z$, giving rise to the joint-contrastive denoising VI (\texttt{JC-Den}). In the interest of space the detailed description of these variants is omitted here.

As the denoising criterion estimates the gradients of ELBO but not the value itself, the denoising approach does not provide a direct way to learn model parameters $\theta$. The denoising method may be best utilised in conjunction with an adversarial algorithm such as a combination of \texttt{PC-Den} and \texttt{PC-Adv}. The denoising method works better early on in the training when $q_\psi$ and $p_\theta$ are very different, and therefore the discrimination task is too easy. Conversely, as $q_\psi$ approaches $p_\theta$, the discriminator can focus its efforts on modelling the residual differences between them rather than trying to model everything about $q_\theta$ in isolation as the denoiser in Algorithm \ref{alg:PCDen} does. \citet{wardefarley2017improving} already used such combination of adversarial training with a denoising criterion for generative modelling. However, the additional nonlinear transformation before denoising introduced in that work breaks the mathematical connections to KL divergence minimisation.

\section{Summary}

To summarise, we have presented two main ways to formulate ELBO in terms of logarithmic density ratios $\priorratio$ and $\jointratio$. We called these prior-contrastive (PC) and joint-contrastive (JC). We have then described two techniques by which these density ratios, or their gradients, can be estimated if the distributions involved are implicit: adversarial methods (\texttt{PC-Adv} and \texttt{JC-Adv}, Algorithms \ref{alg:PCAdv}\&\ref{alg:JCAdv}) directly estimate density ratios via logistic regression, denoising methods (\texttt{DC-Den} and \texttt{JC-Den}, Algorithm \ref{alg:PCDen}) estimate gradients of the log densities via denoising autoencoders. We have mentioned that these methods can be combined, and that such combination may improve convergence.

While all of these algorithms can perform variational inference - fitting the variational parameters $\psi$ - not all of them can perform full variational learning of model parameters $\theta$ if the model itself is implicitly defined. In Section \ref{sec:discussion} we outline an idea based on reverse mode differentiation (RMD) idea by \citet{maclaurin2015gradient} which, giving rise to an algorithm we refer to as \texttt{JC-Adv-RMD}, which can in theory perform fully variational inference and learning in a model where all distributions involved are implicit.

The capabilities of algorithms presented here and in related work are summarised in Table \ref{summary-table}. The adversarial variational Bayes \citep{mescheder2017adversarial} is equivalent to PC-Adv, while ALI \citep{dumoulin2017adversarially} and BiGAN \citep{donahue2017adversarial} are closely related to JC-Adv. \citep{sonderby2016amortised} and \citep{dosovitskiy2016generating} are closely related to PC-Adv, although the former solves a limited special case and the latter uses the Jensen-Shannon formulation and hence is not fully variational.

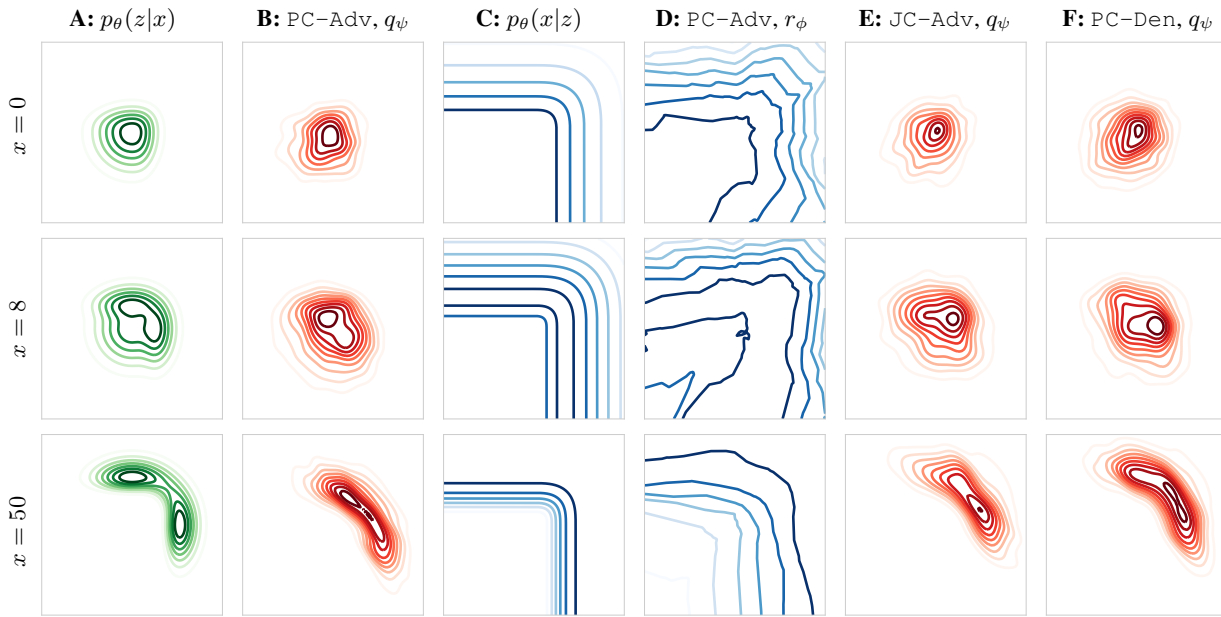
\begin{figure*}[t]
\label{fig:posteriors}
\begin{tikzpicture}
\node (subplots) at (0,0){\includegraphics[width=0.95\textwidth]{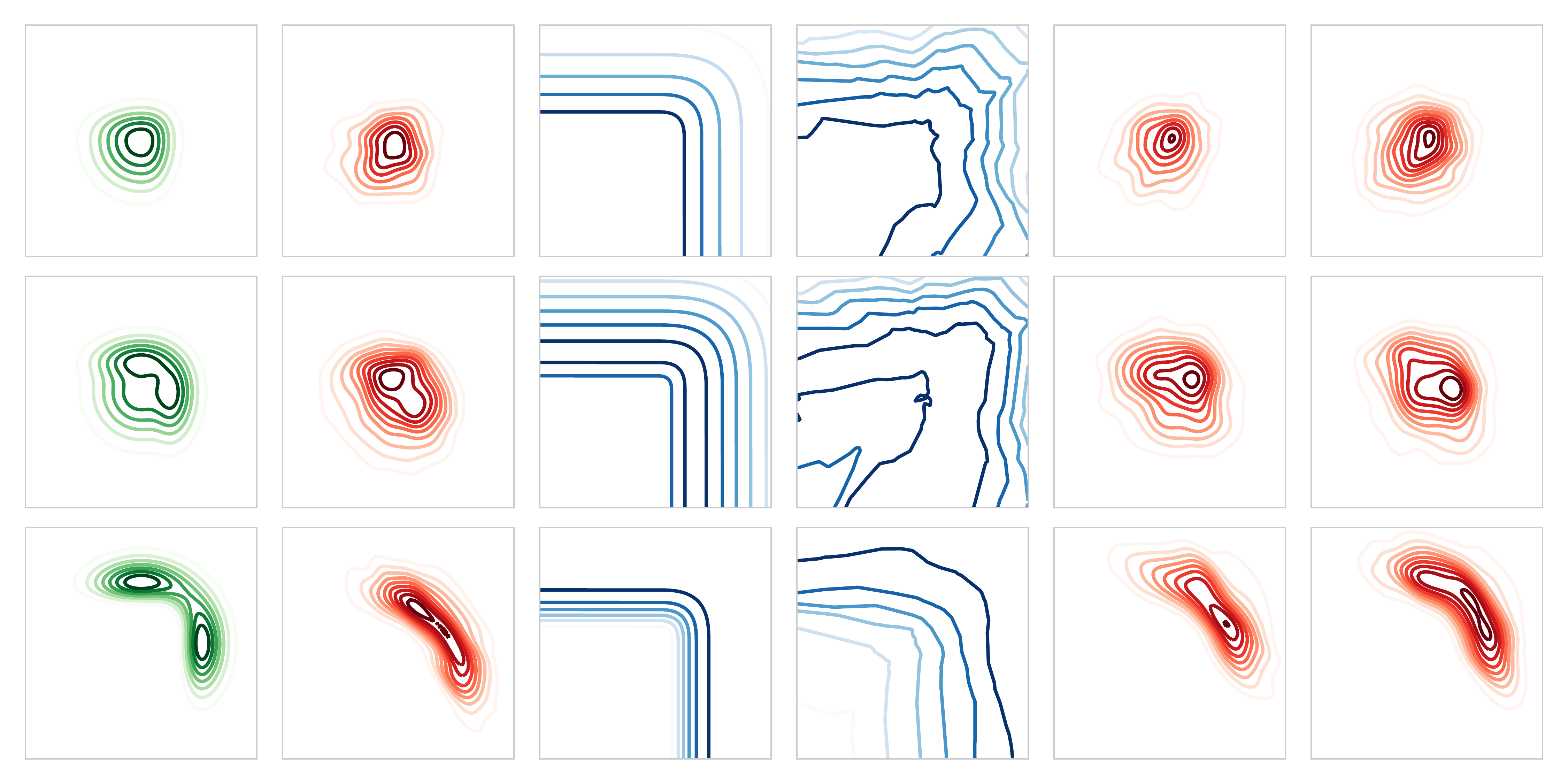}};
\node (xequals0) at (-8.2,2.7)[rotate=90] {\small $x=0$};
\node (xequals8) at (-8.2,0)[rotate=90] {\small $x=8$};
\node (xequals50) at (-8.2,-2.7)[rotate=90] {\small $x=50$};

\node (realposterior) at (-6.8,4.1) {\small \textbf{A:}\ $p_\theta(z\vert x)$};
\node (pcadvpost) at (-4.0,4.1) {\small \textbf{B:}\ \texttt{PC-Adv}, $q_\psi$};
\node (reallikelihood) at (-1.4,4.1) {\small \textbf{C:}\ $p_\theta(x\vert z)$};
\node (pcadvratio) at (1.3,4.1) {\small \textbf{D:}\ \texttt{PC-Adv}, $\priorratio_\phi$};
\node (jcadv) at (4.0,4.1) {\small \textbf{E:}\ \texttt{JC-Adv}, $q_\psi$};
\node (pcden) at (6.7,4.1) {\small \textbf{F:}\ \texttt{PC-Den}, $q_\psi$};
\end{tikzpicture}

\caption{Approximate inference in the ``continuous sprinkler'' toy model from Section \ref{sec:experiments}. Rows correspond to different values of $x$. Column A shows contours of the real posterior $p_\theta(z\vert x)$. With increasing values of $x$ the conditional dependence between $z_1$ and $z_2$ increases. Columns B, E and F show the approximate posterior $q_\psi$ obtained by \texttt{PC-Adv} (Alg.\ \ref{alg:PCAdv}), \texttt{JC-Adv} (Alg.\ \ref{alg:JCAdv}) and \texttt{PC-Den} (Alg.\ \ref{alg:PCDen}), respectively. We could not observe any systematic difference in the algorithms' final estimates. The quality of approximation seems to be predominantly influenced by the choice of architecture used to implement $q_\psi$. Column D shows the final estimate $\priorratio_\phi$ in the \texttt{PC-Adv} algorithm. If the approximate posterior was perfect, $\priorratio_\phi$ should match the likelihood $p_\theta(x\vert z)$, shown in column C, up to an additive constant.}
\end{figure*}

\section{Experiments}\label{sec:experiments}

Several related papers have already demonstrated the success of methods surveyed here on real world datasets, see for example \citep{mescheder2017adversarial,dosovitskiy2016generating} for \texttt{PC-Adv}, \citep{dumoulin2017adversarially,donahue2017adversarial} for \texttt{JC-Adv} and \citep{sonderby2016amortised,wardefarley2017improving} for denoiser-based techniques. Experiments in these papers typically focus on the models' ability to learn $\theta$, and the quality of samples from the learnt generative model $p_\theta(x)$.

As the focus here is on inference rather than learning, the goal of this section is to validate the algorithms' ability to perform inference. To this end, we have devised a simple toy problem loosely based on the ``sprinkle'' example which exhibits \emph{explaining away} \citep{pearl1988embracing}. In our ``continuous sprinkler'' model, two independent scalar hidden variables $z_1$ and $z_2$ are combined nonlinearly to produce a univariate observation $x$:
\begin{align*}
(z_1,z_2) &\sim \mathcal{N}(0, \sigma^2 I_{2\times2})\\
x &\sim \operatorname{EXP}(3 + \max(0,z_1)^3 + \max(0,z_2)^3)
\end{align*}

Although $z_1$ and $z_2$ are a priori independent, the likelihood introduces dependence between the two variables once conditioned on data: either latent variable taking a large value can explain a large observed $x$. This is an example of explaining away which is an important phenomenon in latent variable models that is known to be hard to model with simple, unimodal distributions.

Column A in Figure \ref{fig:posteriors} illustrates the joint posterior density of $z_1$ and $z_2$ for various values of $x$. The subsequent columns show the posterior approximations by \texttt{PC-Adv}, \texttt{JC-Adv} and \texttt{PC-Den}, respectively. $q_\theta$ is implemented as a stochastic generative network where the observation $x$ and Gaussian noise variables $\epsilon$ are fed as input to a multilayer perceptron $g_\psi$. The discriminators and denoisers were implemented as multilayer perceptrons as well. Columns C and D illustrate the limiting behaviour of the discriminator in the \texttt{PC-Adv} algorithm: as $q_\psi$ converges to the true posterior, $\priorratio_\psi$ is expected resemble the likelihood $p_\theta(x\vert z)$ up to an additive constant. In \texttt{JC-Adv} the discriminator eventually converges to the flat $\jointratio_\phi=0$ solution.

\section{Discussion and future work}\label{sec:discussion}

\textbf{Are adversaries really needed?} When using adversarial techniques for VI, we model the distribution of latent variables rather than observations. The distributions we encounter in VI are usually thought of as simpler than the distribution of observed data, so the question arises whether the flexibility of the adversarial framework is really needed.

\textbf{Is the prior-contrastive too much like noise-contrastive?} In the \texttt{PC-Adv} algorithm, the discriminator compares samples from the approximate posterior to the prior, and the prior is often high-dimensional Gaussian noise. Even at convergence, the two distributions the discriminator sees never overlap, and this may slow down training. This can be remedied by observing that as $q_\psi$ converges to the true posterior, the discriminator will converge to the log-likelihood plus constant $\priorratio_{\phi}(x,z)\approx \log p(x| y) + c$. Hence, the task of the discriminator can be made easier by forming an ensemble between a neural network and the actual log-likelihood.

\textbf{Aren't denoising methods imprecise?} The main criticism of denoiser-based methods is that the gradient estimates are imprecise. As \citep{valpola2016denoising} pointed out, the optimal denoising function represents the gradients of the noise-corrupted distribution rather than the original, and in practical cases the noise level $\sigma_n$ may not be small enough for this effect to be negligible. \citet{sonderby2016amortised} observed that denoiser-based methods can not produce results as sharp as adversarial counterparts. Finally, for the outer loop SGD to work consistently, the gradient estimates provided by the inner loop have to form a conservative vector field. While the Bayes-optimal denoiser function satisfies this, it is unclear to what degree this property is preserved when using suboptimal denoisers \citep{im2016conservativeness}. We believe that an alternative approach based on score matching \citep{hyvarinen2005estimation} - a task intimately related to denoising \citep{vincent2011connection} - might overcome both of these issues.

\textbf{How to learn $\theta$?} The focus of this note is on variational inference, which is finding $\psi$. However, it is equally important to think about learning $\theta$. Unfortunately, none of the algorithms presented here allow for fully variational learning of model parameters $\theta$ when $p_\theta(x\vert z)$ is implicit. ALI and BiGAN do provide an algorithm, but as we mentioned, they are not fully variational. We close by highlighting one possible avenue for future work to enable this: differentiating the inner loop of the \texttt{JC-Adv} algorithm via reverse mode differentiation \citep[RMD,][]{maclaurin2015gradient}. To learn $\theta$ via SGD, one only needs an estimate of the gradient $\frac{\partial \jointratio_{\psi,\theta}}{\partial \theta}$. We can't compute $\jointratio_{\psi,\theta},$ only an approximation $\jointratio_\phi$ which is reached via SGD. Each step of SGD depends implicitly on $\theta$. Following \citet{maclaurin2015gradient} we can algorithmically differentiate the SGD algorithm in a memory-efficient way to obtain an estimate of the gradient we need for learning $\theta$. We have not validated this approach experimentally, but included it as \texttt{JC-Adv-RMD} in Table \ref{summary-table}.

\bibliography{main}

\begin{thebibliography}{33}
\providecommand{\natexlab}[1]{#1}
\providecommand{\url}[1]{\texttt{#1}}
\expandafter\ifx\csname urlstyle\endcsname\relax
  \providecommand{\doi}[1]{doi: #1}\else
  \providecommand{\doi}{doi: \begingroup \urlstyle{rm}\Url}\fi

\bibitem[Alain \& Bengio(2014)Alain and Bengio]{alain2014regularized}
Alain, Guillaume and Bengio, Yoshua.
\newblock What regularized auto-encoders learn from the data-generating
  distribution?
\newblock \emph{Journal of Machine Learning Research}, 15\penalty0
  (1):\penalty0 3563--3593, 2014.

\bibitem[Bickel et~al.(2007)Bickel, Br{\"u}ckner, and
  Scheffer]{bickel2007discriminative}
Bickel, Steffen, Br{\"u}ckner, Michael, and Scheffer, Tobias.
\newblock Discriminative learning for differing training and test
  distributions.
\newblock In \emph{Proceedings of the 24th International Conference on Machine
  learning}, pp.\  81--88. ACM, 2007.

\bibitem[Donahue et~al.(2017)Donahue, Krähenb{\:{u}}hl, and
  Darrell]{donahue2017adversarial}
Donahue, Jeff, Krähenb{\:{u}}hl, Philipp, and Darrell, Trevor.
\newblock Adversarially feature learning.
\newblock In \emph{International Conference on Learning Representations}, 2017.
\newblock URL \url{https://arxiv.org/abs/1605.09782}.

\bibitem[Dosovitskiy \& Brox(2016)Dosovitskiy and
  Brox]{dosovitskiy2016generating}
Dosovitskiy, Alexey and Brox, Thomas.
\newblock Generating images with perceptual similarity metrics based on deep
  networks.
\newblock In \emph{Advances in Neural Information Processing Systems}, pp.\
  658--666, 2016.

\bibitem[Dumoulin et~al.(2017)Dumoulin, Belghazi, Poole, Lamb, Arjovsky,
  Mastropietro, and Courville]{dumoulin2017adversarially}
Dumoulin, Vincent, Belghazi, Ishmael, Poole, Ben, Lamb, Alex, Arjovsky, Martin,
  Mastropietro, Olivier, and Courville, Aaron.
\newblock Adversarially learned inference.
\newblock In \emph{International Conference on Learning Representations}, 2017.
\newblock URL \url{https://arxiv.org/abs/1606.00704}.

\bibitem[Goodfellow et~al.(2014)Goodfellow, Pouget-Abadie, Mirza, Xu,
  Warde-Farley, Ozair, Courville, and Bengio]{goodfellow2014generative}
Goodfellow, Ian, Pouget-Abadie, Jean, Mirza, Mehdi, Xu, Bing, Warde-Farley,
  David, Ozair, Sherjil, Courville, Aaron, and Bengio, Yoshua.
\newblock Generative adversarial nets.
\newblock In \emph{Advances in Neural Information Processing Systems}, pp.\
  2672--2680, 2014.

\bibitem[Hyv{\"a}rinen(2005)]{hyvarinen2005estimation}
Hyv{\"a}rinen, Aapo.
\newblock Estimation of non-normalized statistical models by score matching.
\newblock \emph{Journal of Machine Learning Research}, 6\penalty0
  (Apr):\penalty0 695--709, 2005.

\bibitem[Im et~al.(2016)Im, Belghazi, and Memisevic]{im2016conservativeness}
Im, Daniel~Jiwoong, Belghazi, Mohamed~Ishmael, and Memisevic, Roland.
\newblock Conservativeness of untied auto-encoders.
\newblock In \emph{Proceedings of the Thirtieth AAAI Conference on Artificial
  Intelligence}, pp.\  1694--1700. AAAI Press, 2016.

\bibitem[Kanamori et~al.(2009)Kanamori, Hido, and Sugiyama]{kanamori2009least}
Kanamori, Takafumi, Hido, Shohei, and Sugiyama, Masashi.
\newblock A least-squares approach to direct importance estimation.
\newblock \emph{Journal of Machine Learning Research}, 10\penalty0
  (Jul):\penalty0 1391--1445, 2009.

\bibitem[Karaletsos(2016)]{karaletsos2016adversarial}
Karaletsos, Theofanis.
\newblock Adversarial message passing for graphical models.
\newblock \emph{arXiv preprint arXiv:1612.05048}, 2016.
\newblock URL \url{https://arxiv.org/abs/1612.05048}.

\bibitem[Kingma \& Welling(2014)Kingma and Welling]{kingma2015autoencoding}
Kingma, Diderik and Welling, Max.
\newblock Auto-encoding variational {Bayes}.
\newblock In \emph{International Conference on Learning Representations}, 2014.
\newblock URL \url{https://arxiv.org/abs/1312.6114}.

\bibitem[Kingma et~al.(2016)Kingma, Salimans, and Welling]{kingma2016improving}
Kingma, Diederik~P, Salimans, Tim, and Welling, Max.
\newblock Improving variational inference with inverse autoregressive flow.
\newblock \emph{arXiv preprint arXiv:1606.04934}, 2016.

\bibitem[Ledig et~al.(2016)Ledig, Theis, Husz{\'a}r, Caballero, Cunningham,
  Acosta, Aitken, Tejani, Totz, Wang, et~al.]{ledig2016photo}
Ledig, Christian, Theis, Lucas, Husz{\'a}r, Ferenc, Caballero, Jose,
  Cunningham, Andrew, Acosta, Alejandro, Aitken, Andrew, Tejani, Alykhan, Totz,
  Johannes, Wang, Zehan, et~al.
\newblock Photo-realistic single image super-resolution using a generative
  adversarial network.
\newblock \emph{arXiv preprint arXiv:1609.04802}, 2016.

\bibitem[Li \& Turner(2016)Li and Turner]{li2016renyi}
Li, Yingzhen and Turner, Richard~E.
\newblock R{\'e}nyi divergence variational inference.
\newblock In \emph{Advances in Neural Information Processing Systems}, pp.\
  1073--1081, 2016.

\bibitem[Maclaurin et~al.(2015)Maclaurin, Duvenaud, and
  Adams]{maclaurin2015gradient}
Maclaurin, Dougal, Duvenaud, David~K, and Adams, Ryan~P.
\newblock Gradient-based hyperparameter optimization through reversible
  learning.
\newblock In \emph{International Conference on Machine Learning}, pp.\
  2113--2122, 2015.

\bibitem[Makhzani et~al.(2016)Makhzani, Shlens, Jaitly, and
  Goodfellow]{makhzani2016adversarial}
Makhzani, Alireza, Shlens, Jonathon, Jaitly, Navdeep, and Goodfellow, Ian.
\newblock Adversarial autoencoders.
\newblock In \emph{International Conference on Learning Representations}, 2016.
\newblock URL \url{http://arxiv.org/abs/1511.05644}.

\bibitem[Mescheder et~al.(2017)Mescheder, Nowozin, and
  Geiger]{mescheder2017adversarial}
Mescheder, Lars~M., Nowozin, Sebastian, and Geiger, Andreas.
\newblock Adversarial variational {Bayes}: Unifying variational autoencoders
  and generative adversarial networks.
\newblock \emph{CoRR}, abs/1701.04722, 2017.
\newblock URL \url{http://arxiv.org/abs/1701.04722}.

\bibitem[Minka(2001)]{minka2001expectation}
Minka, Thomas~P.
\newblock Expectation propagation for approximate bayesian inference.
\newblock In \emph{Proceedings of the Seventeenth conference on Uncertainty in
  artificial intelligence}, pp.\  362--369. Morgan Kaufmann Publishers Inc.,
  2001.

\bibitem[Mohamed \& Lakshminarayanan(2016)Mohamed and
  Lakshminarayanan]{mohamed2016learning}
Mohamed, Shakir and Lakshminarayanan, Balaji.
\newblock Learning in implicit generative models.
\newblock \emph{arXiv preprint arXiv:1610.03483}, 2016.

\bibitem[Pearl(1988)]{pearl1988embracing}
Pearl, Judea.
\newblock Embracing causality in default reasoning.
\newblock \emph{Artificial Intelligence}, 35\penalty0 (2):\penalty0 259--271,
  1988.

\bibitem[Radford et~al.(2016)Radford, Metz, and
  Chintala]{radford2015unsupervised}
Radford, Alec, Metz, Luke, and Chintala, Soumith.
\newblock Unsupervised representation learning with deep convolutional
  generative adversarial networks.
\newblock In \emph{International Conference on Learning Representations}, 2016.
\newblock URL \url{https://arxiv.org/abs/1511.06434}.

\bibitem[Ranganath et~al.(2016)Ranganath, Tran, Altosaar, and
  Blei]{ranganath2016operator}
Ranganath, Rajesh, Tran, Dustin, Altosaar, Jaan, and Blei, David.
\newblock Operator variational inference.
\newblock In \emph{Advances in Neural Information Processing Systems}, pp.\
  496--504, 2016.

\bibitem[Rezende \& Mohamed(2015)Rezende and Mohamed]{icml2015_rezende15}
Rezende, Danilo and Mohamed, Shakir.
\newblock Variational inference with normalizing flows.
\newblock In Blei, David and Bach, Francis (eds.), \emph{Proceedings of the
  32nd International Conference on Machine Learning (ICML-15)}, pp.\
  1530--1538. JMLR Workshop and Conference Proceedings, 2015.
\newblock URL \url{http://jmlr.org/proceedings/papers/v37/rezende15.pdf}.

\bibitem[Salimans et~al.(2016)Salimans, Goodfellow, Zaremba, Cheung, Radford,
  and Chen]{salimans2016improved}
Salimans, Tim, Goodfellow, Ian, Zaremba, Wojciech, Cheung, Vicki, Radford,
  Alec, and Chen, Xi.
\newblock Improved techniques for training gans.
\newblock In \emph{Advances in Neural Information Processing Systems}, pp.\
  2226--2234, 2016.

\bibitem[S{\o}nderby et~al.(2016)S{\o}nderby, Raiko, Maal{\o}e, S{\o}nderby,
  and Winther]{sonderby2016ladder}
S{\o}nderby, Casper~Kaae, Raiko, Tapani, Maal{\o}e, Lars, S{\o}nderby,
  S{\o}ren~Kaae, and Winther, Ole.
\newblock Ladder variational autoencoders.
\newblock In \emph{Advances in Neural Information Processing Systems}, pp.\
  3738--3746, 2016.

\bibitem[S{\o}nderby et~al.(2017)S{\o}nderby, Caballero, Theis, Shi, and
  Husz{\'a}r]{sonderby2016amortised}
S{\o}nderby, Casper~Kaae, Caballero, Jose, Theis, Lucas, Shi, Wenzhe, and
  Husz{\'a}r, Ferenc.
\newblock Amortised {MAP} inference for image super-resolution.
\newblock In \emph{International Conference on Learning Representations}, 2017.

\bibitem[Sugiyama et~al.(2008)Sugiyama, Nakajima, Kashima, Buenau, and
  Kawanabe]{sugiyama2008kliep}
Sugiyama, Masashi, Nakajima, Shinichi, Kashima, Hisashi, Buenau, Paul~V, and
  Kawanabe, Motoaki.
\newblock Direct importance estimation with model selection and its application
  to covariate shift adaptation.
\newblock In \emph{Advances in neural information processing systems}, pp.\
  1433--1440, 2008.

\bibitem[Uehara et~al.(2016)Uehara, Sato, Suzuki, Nakayama, and
  Matsuo]{uehara2016generative}
Uehara, Masatoshi, Sato, Issei, Suzuki, Masahiro, Nakayama, Kotaro, and Matsuo,
  Yutaka.
\newblock Generative adversarial nets from a density ratio estimation
  perspective.
\newblock \emph{arXiv preprint arXiv:1610.02920}, 2016.

\bibitem[Vajda(2014)]{vajda2014probabilistic}
Vajda, Steven.
\newblock \emph{Probabilistic programming}.
\newblock Academic Press, 2014.

\bibitem[{Valpola et al.}(2016)]{valpola2016denoising}
{Valpola et al.}
\newblock Learning by denoising part 2. connection between data distribution
  and denoising function.
\newblock \url{https://thecuriousaicompany.com/connection-to-g/}, 2016.
\newblock retrieved on 24 February 2017.

\bibitem[Vincent(2011)]{vincent2011connection}
Vincent, Pascal.
\newblock A connection between score matching and denoising autoencoders.
\newblock \emph{Neural computation}, 23\penalty0 (7):\penalty0 1661--1674,
  2011.

\bibitem[Vincent et~al.(2008)Vincent, Larochelle, Bengio, and
  Manzagol]{vincent2008extracting}
Vincent, Pascal, Larochelle, Hugo, Bengio, Yoshua, and Manzagol,
  Pierre-Antoine.
\newblock Extracting and composing robust features with denoising autoencoders.
\newblock In \emph{Proceedings of the 25th International Conference on Machine
  learning}, pp.\  1096--1103. ACM, 2008.

\bibitem[Warde-Farley \& Bengio(2017)Warde-Farley and
  Bengio]{wardefarley2017improving}
Warde-Farley, David and Bengio, Yoshua.
\newblock Improving generative adversarial networks with denoising feature
  matching.
\newblock In \emph{International Conference on Learning Representations}, 2017.
\newblock URL \url{https://openreview.net/forum?id=S1X7nhsxl&noteId=S1X7nhsxl}.

\end{thebibliography}
\bibliographystyle{icml2017}

\appendix
\newpage
\section{Ignoring implicit dependence on $\psi$}\label{sec:proof_implicit_dependency}

Proof of Eqn.\ \eqref{eqn:implicit_dependency_ignored}:

\begin{align}
&\left. \frac{\partial}{\partial \psi} \kl{q_\psi}{p_\theta} \right\vert_{\psi = \psi_0} = \frac{\partial}{\partial \psi} \left.\expect{z \sim q_\psi} \priorratio_{\theta, \psi}(z) \right\vert_{\psi = \psi_0} \notag\\
 &= \left. \frac{\partial}{\partial \psi} \expect{z \sim q_\psi}\log\frac{q_{\psi_0}(z)}{p(z)} \right\vert_{\psi = \psi_0} + \left. \frac{\partial}{\partial \psi} \kl{q_\psi}{q_{\psi_0}} \right\vert_{\psi = \psi_0} \notag\\
 &= \left. \frac{\partial}{\partial \psi} \expect{z \sim q_\psi}\log\frac{q_{\psi_0}(z)}{p(z)} \right\vert_{\psi = \psi_0} \notag\\
  &= \left. \frac{\partial}{\partial \psi} \expect{z \sim q_\psi} \priorratio_{\theta, \psi_0}(z) \right\vert_{\psi = \psi_0},
\end{align}
where the third line is obtained by noting that $\psi_0$ is a local minimum of $\kl{q_\psi}{q_{\psi_0}}$, hence the second term in the second line is $0$.

\end{document}